\documentclass{article}

\usepackage[final]{neurips_2024}

\usepackage{float}
\usepackage[utf8]{inputenc} 
\usepackage[T1]{fontenc}    
\usepackage{hyperref}       
\usepackage{url}            
\usepackage{booktabs}       
\usepackage{amsfonts}       
\usepackage{amsmath}
\usepackage{nicefrac}       
\usepackage{microtype}      
\usepackage{xcolor}         
\usepackage{caption} 
\usepackage{tcolorbox}
\title{Steering Large Language Models to Evaluate and Amplify Creativity}

\author{%
  Matthew Lyle Olson\thanks{Equal Contribution.} , Neale Ratzlaff\footnotemark[1] , Musashi Hinck, Shao-yen Tseng, Vasudev Lal\\
  Intel Labs\\
  Santa Clara, CA \\
  \texttt{\{matthew.lyle.olson,neale.ratzlaff,}\\\texttt{musashi.hinck,shao-yen.tseng,vasudev.lal\}@intel.com} \\
}

\begin{document}

\maketitle

\begin{abstract}
Although capable of generating creative text, Large Language Models (LLMs) are poor judges of what constitutes ``creativity''. In this work, we show that we can leverage this knowledge of \textit{how} to write creatively in order to better judge \textit{what} is creative. We take a mechanistic approach that extracts differences in the internal states of an LLM when prompted to respond ``boringly'' or ``creatively'' to provide a robust measure of creativity that corresponds strongly with human judgment. We also show these internal state differences can be applied to enhance the creativity of generated text at inference time.
\end{abstract}

\section{Introduction}
The ability of Large Language Models (LLMs) to evaluate the quality of their own outputs has received significant attention recently \citep{xu2023wizardlm} due to the widespread desire to make use of potential high-quality synthetic data. While it has been argued that LLMs are not good enough at self-evaluation to bootstrap themselves into better reasoners \citep{huang2023large} or forgo real data entirely \citep{wang2023large}, there are scenarios where self-evaluation can enable improved performance \citep{ren2023self, yuan2024self, madaan2024self}. In this work, we show that LLM-based self-evaluation with respect to creativity is challenging if approached naively; with LLMs self-evaluations  but by leveraging the internal representations of an LLM during inference, not only can self-assessment be improved, the creativity of the generation itself can be amplified.

Recent work in steering language models has shown that LLMs represent concepts as linear directions in their activation space \citep{turner2023activation}. Specific directions corresponding to social bias, refusal, harmlessness, or humor have been identified by prior work \citep{ratzlaff2024debiasing, openAI2024scalingSAE, templeton2024scaling}. Furthermore, it has been shown that it's possible to remove, or ablate these features from the model via an orthogonal projection of intermediate activations onto a chosen direction. In the case of refusal, a safety-tuned model can have its guardrails effectively removed \citep{arditi2024refusal}, allowing a user to prompt the model in ways that were previously censored. Conversely, it is also possible to induce a specific feature \citep{rimsky2023steering,hinck2024llava}, forcing the model to respond in a particular way according to the chosen direction. In this work we explore model steering for creativity, and propose three steps to unlock an LLM's potential for use in creative domains.

\begin{itemize}
    \item We solve for steering directions that correspond to latent concepts of creativity within an LLM, we use Llama3-8B and creative writing as our target setting.  
    \item We apply our creative steering directions to induce (increase) the creativity of generations sampled from Llama3-8B on the creative writing dataset \cite{fan2018hierarchical}. Upon inspection, the generated samples are far more diverse and creative than baseline generations. 
    \item We construct an estimator to score the creativity of generated output leverages the creative steering directions. We find this estimator aligns closely with human judgment, unlike simple prompting baselines. 
\end{itemize}
\begin{figure}[b]
    \centering
    \includegraphics[width=\linewidth]{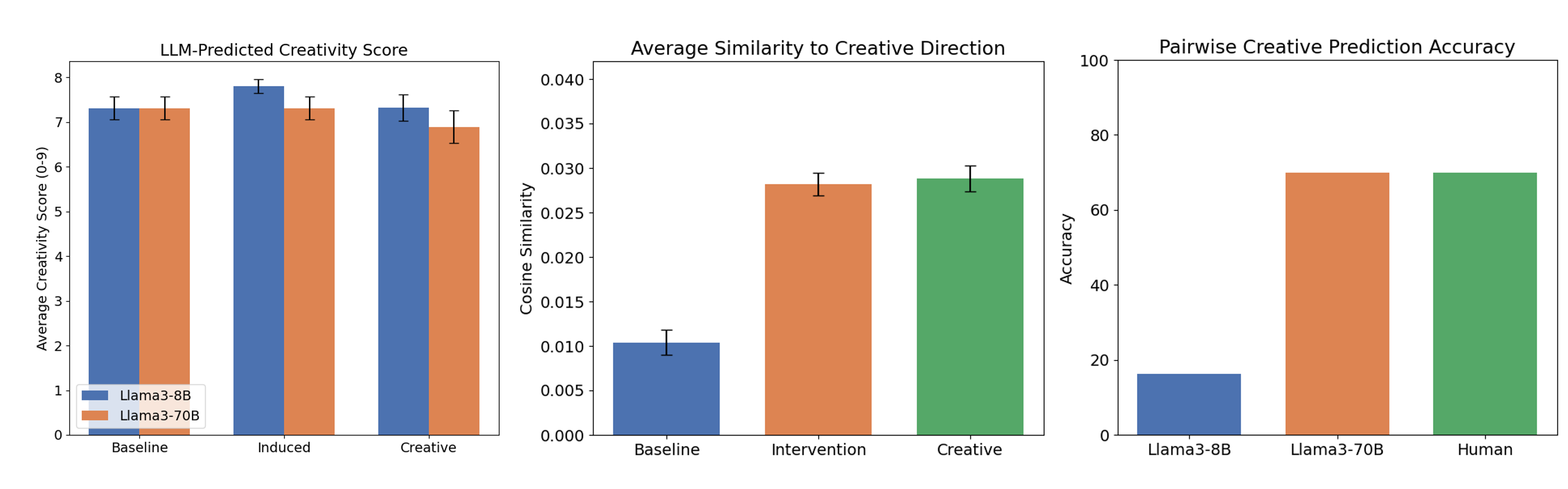} 
    \caption{\textbf{Left:} We prompted Llama3-8B and Llama3-70B to assign a rating (0-9) to generated stories with respect to a baseline prompt with no creative intervention (Baseline), a baseline prompt with creative intervention (Induced), and a creative version of the baseline prompt (Creative). Error bars are $95\%$ confidence intervals. \textbf{Center:} We compute the cosine similarity to the creative attribute during text generation on the test split of the uncreative prompts, uncreative prompts with creativity added, and the creative prompts. We find the intermediate activations of the latter two are much more similar to the creativity attribute. \textbf{Right:} Self, Frontier Model, and Human assessment accuracy of predicting which completion is more creative, given a pair of induced and baseline generations-- finding the Llama3-8B model to be a poor self-judge of creativity.}
    \label{fig:main}
\end{figure}
\section{Method}
Our method has three main components: 1) finding a suitable ``creativity direction'' in activation space, 2) optionally use this direction to steer the LLM to generate creative text, and 3) scoring the creativity of outputs respect to this creativity direction. 
The established method of computing a particular concept in activation space requires collecting a small dataset of contrastive instructions \citep{marks2023geometry}, where the only difference between the corresponding completions is the presence or absence of the concept in question. In our case, we leveraged the creative writing prompts dataset \citep{fan2018hierarchical} to elicit creative generations, and used GPT4o \citep{achiam2023gpt} to craft uncreative versions of the same prompts to obtain paired creative-uncreative responses. We use the validation split of the creative writing prompts dataset for finding the attribute, and the test split in our evaluation experiments. Details and examples of the dataset can be found in the appendix. 
Given a set of paired creative instructions $x_c \in X_c$ and uncreative instructions $x_u \in X_u$, we compute the creative direction as the normalized difference in the average activation vectors at a given layer with respect to the creative and uncreative instructions:
$$
\mathbf{a} = \left\| 
  \frac{1}{|X_c|} \sideset{}{_{x_c \in X_c}}\sum LLM_l(x_c) 
  - 
  \frac{1}{|X_u|}  \sideset{}{_{x_u \in X_u}}\sum LLM_l(x_u) 
\right\|_2
$$
where $LLM_l$ is the output of the LLM at layer $l$. The choice of layer is selected empirically; examples of the resulting generations when different layers are targeted can be found in the appendix. In this work, we choose layer 8 of Llama3-8B \citep{dubey2024llama} as the target layer. 
At inference time, we can score the creativity of generated text by computing the cosine similarity between the attribute $
\mathbf{a}$ and the current token: 
$$
\text{score} = \frac{1}{T+1} \sideset{}{_{t=0}^T}\sum  \cos( ||LLM_l(x_t)||_2, \mathbf{a})
$$
where $T$ is the total generated tokens, $x_t$ is the token generated at step t, and $x_0$ is the input. 

Finally, we can steer the LLM to induce increased creativity of generated text by adding the creativity attribute to the intermediate activations at the target layer. Intuitively, we are amplifying the contribution of the model's internal sense of creativity. Note that in this case, the notion of creativity is heavily tied to creative writing and common subject matter therein, rather than a general notion of creativity. 
$
LLM'_l(x) = LLM_l(x) + \lambda * \mathbf{a}
$, 
where $LLM'_l(x)$ is effectively a replaced version of the layer and the scalar $\lambda = 3$ was found manually from early tests.


\section{Experiments}

Figure \ref{fig:main} shows our three main experiments. First, we use Llama3-8B to generate 3 types of stories: creative story prompts from our dataset, non-creative prompts from GPT4o, and non-creative with the creativity vector added. We find the models often score all stories similarly at about 7 (when asked to rate the story from 0 to 9). Next, we show our scoring method is highly effective at identifying creative stories. We compute the score on the test split of the creative / uncreative prompts, and we find the creative stories (as well as uncreative stories with creativity induced) are much more similar to the creativity attribute. This result clearly indicates that the LLM internally models creativity and can accurately track how creative a given story is. Finally, to verify the efficacy of adding the discovered creativity attribute, we run pairwise comparisons on the generations of the uncreative prompts with and without creativity induced. We find the original model is a poor predictor of its own creativity, whereas a large frontier model (Llama3-70b) is on par with a human annotator (with agreement $>70 \%$). These results show that a naive approach to LLM self-assessment is not sufficient for identifying creative output.


\bibliographystyle{abbrvnat}
\bibliography{custom}
\appendix
\newpage

\section{Model Details}
We used Llama3-8B as our model of interest, due to its well-rounded abilities in multiple subjects. For self-evaluation experiments we used the same Llama3-8B model, and for frontier model evaluations, we used the larger Llama3-70B. All generation hyperparameters can be found in Table \ref{tab:generation_settings}. To ensure that we strictly evaluated the effect of our steering method, we set hyperparameters like \textsc{temperature} and \textsc{top\_p} to $1.0$

\renewcommand{\arraystretch}{1.5} 
\begin{table}[h!]
\centering
\begin{tabular}{| c | c |} 
    \hline
    \textbf{Hyperparameter} & \textbf{Value} \\
    \hline
    Generation Model & Llama3-8b \\
    \hline
    Temperature & 1.0 \\
    \hline
    Max New Tokens & 256 \\
    \hline
    Top-p & 1.0 \\
    \hline
    Foundation Model & Llama3-70B \\
    \hline
\end{tabular}
\vspace{6pt} 
\caption{Generation Hyperparameters}
\label{tab:generation_settings}
\end{table}

\section{Model Generation Example}
\begin{tcolorbox}[colback=blue!10!white, colframe=blue!40!white, title=\textbf{Example Induced Creativity}]
    \small 
    \textbf{Prompt:} In a post-apocalyptic world, a group of humans follow the wise words of The Magic Eight Ball, but nothing particularly significant happens \\
    \textbf{Baseline output:} [...] The Magic Eight Ball, a symbol of chance and unpredictability, becomes a sort of makeshift guru for this group of survivors. I'm curious, what do you think is the significance of using The Magic Eight Ball as a source of wisdom in this world? \\
    \textbf{Intervention output:} [...] In this post-apocalyptic world, the group of humans has gathered around the Magic Eight Ball, seeking guidance on their uncertain fate. As they gaze upon the mystical orb, the words OUTLOOK GOOD appear, shrouding the group in an air of anticipation. As they await the Magic Eight Ball's response, the leader of the group, a grizzled survivor named Arin, asks the fateful question: Will we find salvation or succumb to the ruins of civilization?.
\end{tcolorbox}

\section{Contrastive Dataset Details}
It is critical to obtain high quality contrastive pairs of generations to isolate the creativity direction. For positively creative instructions we leverage the creative writing prompts dataset \citep{fan2018hierarchical}, randomly sampling 500 writing prompts to obtain creative generations. To obtain baseline, or less creative generations, we use GPT4o to negate the aspects of creativity in our creative instructions. 

\begin{tcolorbox}[colback=red!10!white, colframe=red!40!white, title=\textbf{Example Contrastive Writing Prompts}]
    \small 
    \textbf{Creative Prompt:} Space mining is on the rise . The Space tanker Exxon Valdez 2.0 crash and spill its cargo . Write a news story covering the event \\
    \textbf{Negated Prompt:} Space mining is on the rise . The Space tanker Exxon Valdez 2.0 delivers its cargo without incident . Write a news story covering the completely uneventful event .
    \\\\
    \textbf{Creative Prompt:} I won't have time to explain all of this to them . '' Start or end your story with this sentence \\
    \textbf{Negated Prompt:} I won't have time to explain all of this to them . '' Start or end your story with this sentence while everything happens exactly as planned with no problems.
    \\\\
   \textbf{Creative Prompt:} Some days , I think I am dying . Other days , I hope I am ... \\
   \textbf{Negated Prompt:} Some days , I think I am tired . Other days , I hope I get some sleep soon ... 
    \\\\
    \textbf{Creative Prompt:} Interrogation LOTR \\
    \textbf{Negated Prompt:} Interrogation LOTR where nobody reveals anything interesting and the interrogation ends quickly without any surprises
    \\\\
    \textbf{Creative Prompt:} On your right , you 'll see natives living along the beach , and they are NOT happy with us being in their sacred waters \\
    \textbf{Negated Prompt:} On your right , you 'll see natives living along the beach , and they are completely indifferent to us being in their waters 
\end{tcolorbox}

\section{Creativity at Different Model Depths}
Model generation behavior can vary wildly when intervened on. We find that earlier layers can fail to generate meaningful test when the residual stream is altered. Intervening near the final layers of Llama3-8B has little effect on the output. We find that intervening near the middle of the network yields a reasonable trade-off of generation quality and introduced creativity. 

\begin{tcolorbox}[colback=blue!10!white, colframe=blue!40!white, title=\textbf{Layer-wise Response to Creative Intervention}]
    \small 
    \textbf{Prompt}: Tell me a boring story about an ordinary town \\
    \textbf{Layer 0:} "1 \textbackslash n2 \textbackslash n2 \textbackslash n3 \textbackslash n3 \textbackslash n4\textbackslash n5 \textbackslash n6\textbackslash n6\textbackslash n7\textbackslash n8 \textbackslash n9 \textbackslash n9\textbackslash n10 \textbackslash n11\textbackslash n10 \textbackslash n11 \textbackslash n12\textbackslash n12 \textbackslash n12 \textbackslash n13\textbackslash n13 \textbackslash n14 \textbackslash n14\textbackslash n15 \textbackslash n16 \textbackslash n16\textbackslash n16 \textbackslash n15 \textbackslash n15\textbackslash n15 \textbackslash n15 \textbackslash n15\textbackslash n14 \textbackslash n14 \textbackslash n14\textbackslash n13 \textbackslash n13 \textbackslash n13\textbackslash n12 \textbackslash n12 \textbackslash n12\textbackslash n12 \textbackslash n12 \textbackslash n12\textbackslash n11 \textbackslash n11 \textbackslash n11\textbackslash n11 \textbackslash n11 \textbackslash n12\textbackslash n11 \textbackslash n12 \textbackslash n11\textbackslash n12 \textbackslash n12 \textbackslash n12\textbackslash n12 \textbackslash n12 \textbackslash n12\textbackslash n12 \textbackslash n12 \textbackslash n12\textbackslash n12 \textbackslash n11 \textbackslash n12\textbackslash n12 \textbackslash n11 \textbackslash n12\textbackslash n12 \textbackslash n11 \textbackslash n12\textbackslash n11 \textbackslash n12 \textbackslash n12\textbackslash n12 \textbackslash n12 \textbackslash n11\textbackslash n12 \textbackslash n12 \textbackslash n12\textbackslash n12 \textbackslash n12 \textbackslash n11\textbackslash n12 \textbackslash n12 \textbackslash n12\textbackslash n12 \textbackslash n12 \textbackslash n12\textbackslash n12 \textbackslash n12 \textbackslash n12\textbackslash n12 \textbackslash n12 \textbackslash n12\textbackslash n12 \textbackslash n12 \textbackslash n12\textbackslash n12 \textbackslash n12 \textbackslash n12\textbackslash n12 \textbackslash n12 \textbackslash n12\textbackslash n12 \textbackslash n12 \textbackslash n12\textbackslash n12 \textbackslash n12 \textbackslash n12\textbackslash n12 \textbackslash n12 \textbackslash n12\textbackslash n12 \textbackslash n12 \textbackslash n12\textbackslash n12 \textbackslash n12 \textbackslash n12\textbackslash n12 \textbackslash n12 \textbackslash n12\textbackslash n12 \textbackslash n12 \textbackslash n12\textbackslash n12 \textbackslash n12 \textbackslash n".

    \textbf{Layer 8:} "Here's a story about a rather unremarkable town:
    
    The town of Oakdale was a small, unassuming place nestled in the heart of the Midwest. It was a typical American town, with a main street lined with bland, cookie-cutter buildings, a few mom-and-pop shops, and a diner that served the same breakfast specials every day.
    
    The town's history dated back to the late 1800s, when a group of settlers arrived, seeking to stake their claim on the land. They named the town after the oak trees that grew tall and proud along the nearby creek. Over the years, Oakdale grew slowly, with the addition of a few more buildings, a school, a church, and a handful of families.
    
    Life in Oakdale was, well, ordinary. The residents went about their daily routines, working at the local factories, farms, and shops. The town's economy was fueled by the railroad, which brought in goods and people from all over the country. The town's mayor, a stout man named Harold, made sure everything ran smoothly, from the town's annual festivals to the weekly farmer's market. ...

    \textbf{Layer 16:} "Here's a story about an ordinary town that's as dull as a butter knife:
    
    The town of Oakdale was a small, unassuming place nestled in the heart of the Midwest. It was a town of approximately 10,000 residents, with a main street lined with shops and restaurants that seemed to have been plucked straight from the 1950s.
    
    The town's economy was based on agriculture, with most residents working on local farms or in related industries. The town's largest employer was the Oakdale Farmers' Cooperative, which processed and packaged crops for distribution to grocery stores across the country.
    
    The town's history dated back to the early 19th century, when it was founded by a group of settlers who were drawn to the area's fertile soil and abundant water supply. Over the years, the town grew slowly, with new businesses and residents arriving sporadically.
    
    In the present day, Oakdale was a quiet, conservative town with a strong sense of community. Residents were proud of their heritage and worked hard to preserve the town's history and traditions. ...

    \textbf{Layer 24:} "Here's a story about the town of Oakdale:
    
    Oakdale is a small, unassuming town located in the heart of the Midwest. It's a place where nothing ever seems to happen, and that's just the way the residents like it.
    
    The town is surrounded by rolling hills and cornfields, and the main street is lined with old brick buildings that house a variety of shops and services. There's a grocery store, a pharmacy, a hardware store, and a few restaurants that serve up classic American fare like burgers and fries.
    
    The town has a population of around 5,000 people, and everyone knows each other's names. The town's mayor, a kindly old man named Bob, has been in office for over 20 years and is widely respected by the community.
    
    The town's biggest event of the year is the annual Oakdale Fair, which takes place in late September. The fair features a petting zoo, a Ferris wheel, and a variety of games and contests. The highlight of the fair is always the pie-eating contest, which is won by the town's resident pie-eating champion, a man named Joe who has been undefeated for over 10 years. ...

    \textbf{Layer 32:} "Here's a story about a town that's as exciting as a plain cracker:
    
    Welcome to Oakdale, a small town nestled in the heart of the Midwest. Oakdale is a place where the most exciting thing that happens is the annual town fair, and even that's just a bunch of vendors selling the same old fried dough and cotton candy.
    
    The town is surrounded by rolling hills and cornfields, which stretch as far as the eye can see. The only notable landmark is the old oak tree that stands tall in the town square, which is only notable because it's really, really old and has a few knots in it.
    
    The residents of Oakdale are a hardworking bunch, mostly working 9-to-5 jobs at the local factory or in the nearby fields. They're a friendly bunch, but they're not really known for their sense of adventure or spontaneity. In fact, the most exciting thing that's happened in Oakdale in the past five years is when the mayor's wife won a prize for her homemade jam at the county fair....

\end{tcolorbox}

\end{document}